\documentclass[10pt,journal,compsoc]{IEEEtran}

\usepackage{times,amsmath,epsfig}
\usepackage{xspace}
\usepackage{booktabs}
\usepackage{tabularx}
\usepackage{listings}
\usepackage{graphicx}
\usepackage{url}
\usepackage{subfigure}
\usepackage{multicol}
\usepackage{wrapfig}
\usepackage{bbding}
\usepackage{amsmath}
\usepackage{amssymb}
\usepackage{algorithm}
\usepackage{algorithmicx}
\algrenewcommand\alglinenumber[1]{\scriptsize #1:}
\usepackage{float}
\usepackage{algpseudocode}
\usepackage{color}
\usepackage{setspace}
\usepackage{wrapfig}
\usepackage{pgfplots}
\usepackage{tikz}
\def\IEEEbibitemsep{0pt plus .5pt}

\begin{document}
%
\title{A Deep Reinforcement Learning Approach for Composing Moving IoT Services}

\author{%
{Azadeh Ghari Neiat, Athman Bouguettaya, ~\IEEEmembership{Fellow,~IEEE}, Mohammed Bahutair}
\thanks{Azadeh~Ghari~Neiat is with the School of Information Technology, Deakin University, Geelong, Australia,  Email:azadeh.gharineiat@deakin.edu.au}
\thanks{Athman~Bouguettaya and Mohammed Bahutair are with the School of Computer Science, University of Sydney, Sydney, Australia. Email: \{athman.bouguettaya, mbah6158\}@sydney.edu.au}
}

\markboth{IEEE Transactions on Services Computing}%
{Ghari Neiat \MakeLowercase{\textit{et al.}}: A Deep Reinforcement Learning Approach for Composing Moving IoT Services}

\IEEEcompsoctitleabstractindextext{%

\begin{abstract}

We develop a novel framework for efficiently and effectively discovering crowdsourced services that \emph{move} in close proximity to a user over a period of time. We introduce a moving crowdsourced service model which is modelled as a moving region. We propose a deep reinforcement learning-based composition approach to select and compose moving IoT services considering quality parameters. Additionally, we develop a parallel flock-based service discovery algorithm as a ground-truth to measure the accuracy of the proposed approach. The experiments on two real-world datasets verify the effectiveness and efficiency of the deep reinforcement learning-based approach. 

\end{abstract}

}

\maketitle 

\IEEEdisplaynotcompsoctitleabstractindextext
\IEEEpeerreviewmaketitle 
\setlength{\textfloatsep}{0pt}

\section{Introduction}\label{sec:introduction}

\IEEEPARstart{T}he crowdsharing economy is an emerging dynamic ecosystem where people create new on-demand services through sharing or exchanging resources to achieve mutually beneficial goals \cite{nekaj2017}. This new type of economy has the potential to be applied in a diverse range of sectors, including tourism and hospitality, labor and service platforms, mobility and logistics \cite{Taeihagh2017}. Two well-known examples of are Uber \footnote{www.uber.com} and Airbnb \footnote{www.airbnb.com}. The foundations of this emerging economy are anchored in \textit{crowdsourcing and crowdsharing} \cite{howe2006rise}.
In crowdsourcing environments, crowds need a medium to interact and produce results. A wide variety of devices are used to facilitate certain types of crowdsourcing \cite{burke2006participatory, kazemi2012geocrowd}. In particular, Internet of Things (IoT) devices are usually used to allow the crowd to provide and use {\em services}. We define \textit{services} as an abstraction that transforms IoT data into {\em actionable information} \cite{bouguettaya2017service}. In that respect and more formally, a \textit{service} is defined by its functional and non-functional attributes. A functional attribute is usually defined as {\em what} a service provides, i.e., the purpose of the service. Non-functional attributes are {\em qualities} attached to service provisioning, i.e., Quality of Service (QoS). For example, a functional attribute of an airline service is {\em reservation}. A non-functional attribute of this service is the {\em price} which would be attached to the reservation.


\emph{A moving crowdsourced IoT service} is a service provided by an IoT device \emph{moving} in \emph{time}, \emph{space}, or both. An example of such services is a WiFi hotspot provided by a person through their smartphone. This type of crowdsourced IoT services is characterised by their {\em spatio-temporal} aspects. The \textit{spatio-temporal} properties refer to the location/space and time/period in which crowdsourced services are provisioned and consumed. We use \textit{moving IoT service} and \textit{mobile IoT service} interchangeably.

{\em Mobility} is an important and intrinsic part of the non-functional aspect of crowdsourced services. The mobility of IoT devices provides opportunities to dynamically extend service coverage to larger spaces and at flexible times. Mobility, however, presents key challenges in terms of qualitative factors (e.g., availability) if the aim is to provide users with the best quality of experience. In this respect, we focus on the spatio-temporal aspects as key parameters to query moving crowdsourced services.

\begin{figure}[t]
  \centerline{\psfig{figure=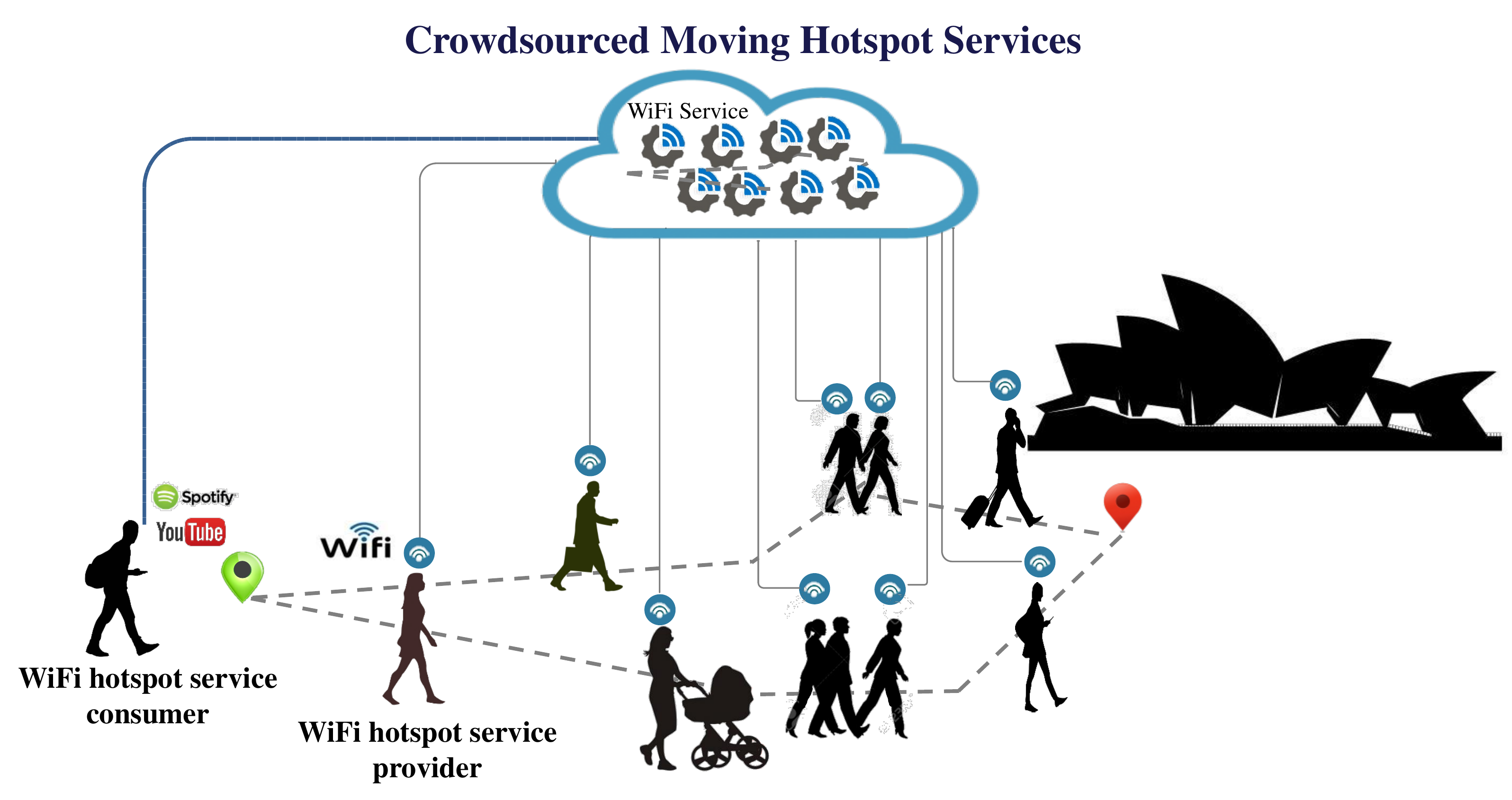,scale=0.28}}
  \caption{WiFi Hotspot Sharing Scenario.}
  \label{fig:scenario} 
  \end{figure}

We identify two key types of crowdsourced services with regard to spatial location: \textit{fixed} and \textit{moving}.
A fixed crowdsourced service refers to services which are permanent in \textit{space} during the time period of the service provisioning. For example, Diana may share her WiFi hotspot while she is sitting at a coffee shop. In contrast, moving crowdsourced  service is not tied to any specific location at any point in time. For example, Diana may share her WiFi hotspot as she moves from one location to another when strolling in the city. More specifically, we assume that a \emph{fixed} hotspot service will remain available (in terms of time and location) when selected by a service consumer and throughout the provisioning of the service. However, for \emph{moving} hotspot services, the availability and location can change during service provisioning. Additionally, crowdsourced services have another dimension, i.e., they may also be \textit{deterministic} or \textit{non-deterministic}. A crowdsourced IoT service is \textit{Deterministic} if \textit{time period} and availability at a certain location are {\em known in advance}. A crowdsourced IoT service is \textit{Non-deterministic} if time period and location availability are {\em not known in advance}. 



There has been a large body of research on service selection and composition in mobile environment \cite{badidi2019personalized,peng2018crowdservice,dengmobility,pramanik2020mobility}. Most existing methods focus on service selection and composition based on Quality of Service (QoS) parameters \cite{deng2016mobilitye} and energy consumption \cite{zhou2018energy,sun2018energy}. They do not take into account the mobility of a user or service provider. There are only a few works that address the mobility-aware service selection problem \cite{wu2019mobility, lakhdari2020composing}. However, these works do not consider the composition of mobile IoT services while both IoT service provider and consumer are moving together.

We identify the following research challenges. The first challenge is \textbf{connectivity} which is an intrinsic part of moving crowdsourced service discovery. \textit{A moving service should stay connected} with a service provider, i.e., be within connectivity proximity of the provider. This requires determining co-movement patterns. We propose a parallel flock-based service discovery to find co-movement services. The parallel flock-based approach is based on a spatio-temporal MapReduce to efficiently find flock patterns \cite{gudmundsson2006computing}. We first apply a \emph{temporal map} step to prune sub-trajectories of moving services with regards to a user trajectory. We then deploy a \emph{spatial map} step to filter candidate moving services which are located within a circular region of a user trajectory. The second challenge is \textbf{service continuity} to connect to the next moving service as an IoT service provider and a user is not necessarily sharing all their route. Therefore, \textit{an effective composition approach is required to select an optimal set of available moving crowdsourced services which ensure the service continuity.}
Most existing trajectory similarity joins approaches are time-interval based \cite{vernica2010efficient,zhang2012efficient} and their methods are not applicable to continuous temporal matching as they retrieve approximate results. Little work \cite{fang2016scalable} addresses the issue of continuous nearest neighbour joins on big trajectory datasets. The key difference with our approach is the need to select continuous sub-trajectories considering a range of QoS parameters. We propose a Deep Reinforcement Learning-based composition algorithm to find and compose valid candidate moving services which overlap with the user trajectory.
Third challenge is \textbf{indexing}. Existing co-movement discovery methods usually rely on \textit{centralized index methods} like R-tree. \cite{fan2016general} shows that the performance of the existing co-movement discovery methods like flock \cite{gudmundsson2006computing}, convoy \cite{jeung2008discovery}, swarm \cite{li2010swarm}, group \cite{ wang2006efficient} and platoon \cite{li2015efficient} degrade dramatically as the dataset scales up. As a result, creating and maintaining an index in parallel computing like MapReduce cannot be effective \cite{fan2016general}. To address this we propose a pruning approach that aims at diverting the algorithm from selecting invalid services without the need to index them.

The contribution of the paper is summarized as follows:
\begin{itemize}
    \item We propose a spatio-temporal model for \emph{moving crowdsourced IoT services}. Our previous work in \cite{neiat2017crowdsourced} and \cite{neiat2015spatio} proposed a selection and composition model for \emph{fixed} crowdsourced services based on spatio-temporal features. We also proposed a temporal non-deterministic service discovery approach. In this work, we focus on \emph{deterministic moving services}.
    \item We design a novel solution based on deep reinforcement learning to support effective discovery and composition without using an index. 
    
    \item We devise a ground-truth, called \emph{parallel flock-based moving crowdsourced service discovery}, using Apache Spark. This is used to measure the accuracy of the proposed discovery model in terms of service discovery in each timestep. The algorithm is based on a spatio-temporal MapReduce. After the spatio-temporal filtering is conducted, we retrieve the valid candidates and feed them to our deep reinforcement learning-based composition approach.
    \item We conduct a set of extensive experiments on two real datasets. The results show the efficiency and effectiveness of the approach in terms of accuracy, learning speed and scalability in comparison with the ground-truth.
\end{itemize}

The rest of the paper is organized as follows: Section 2 surveys related work. Section 3 introduces our system model and states the problem of crowdsourced moving service composition. Section 4 presents the deep reinforcement learning-based composition algorithm. Section 5 provides the ground-truth approach. Section 6 reports our experimental results. Section 7 concludes the related work and highlights future work.


\subsection*{Motivation Scenario}

\textit{Scenario 1:} WiFi tethering may be a connection option when free public WiFi is not available or effective due to low connection speed and limited capacity. In WiFi tethering, the crowd can switch on their IoT devices' WiFi hotspot and share their data balance to other devices for some rewards. As a result, WiFi hotspot can be crowdsourced. IoT devices can be anything the crowd has and is connected to the Internet. Generally, IoT devices can be divided into two main categories: fixed and mobile. Fixed IoT devices are referred to devices that typically do not move, e.g., smart fridge, smart TV, etc. Mobile devices are devices that are inherently made to be carried by people, such as smartwatches, and smartphones. Wearables are a subset of mobile IoT devices. They are special IoT devices that are meant to be worn (e.g., smart shoes and smart shirts). It is predicted that wearables in the near future would become ubiquitous \cite{seneviratne2017survey}.

We consider WiFi hotspot sharing as one of the representatives of crowdsourced IoT services. For instance, it can be leveraged to offer a range of crowdsourced services such as WiFi-coverage travel planning for cost-effective media streaming. In this regard, crowdsourced WiFi hotspot services are provided by smartphones which are moving in space and time through a mobile application. For example, Open Garden\footnote{https://opengarden.com} created an application that lets users share and consume the Internet among each other. They use a monetary incentive, where providers set the price per MB and consumers pay according to their usage.

The trajectories of the services are deterministic. Deterministic trajectories refer to the a-priori knowledge of the trajectories. It is assumed that there is a platform that incentivizes WiFi hotspot providers to  move to specified areas and share their resources \cite{neiat2019incentive}. Therefore, the WiFi hotspot providers are assumed to follow certain trajectories that have been assigned. In particular, we focus on the trajectories of \textit{pedestrians}.  We also assume that WiFi hotspot services will overlay digital maps. We propose to reformulate the research problem of \textit{moving crowdsourced IoT service selection and composition as finding the optimal composition of WiFi hotspot moving services which provide the best quality of experience  to fulfill users' specific requirements/expectations (e.g., watching online videos and signal strength)} (Fig. \ref{fig:scenario}).

 
\textit{Scenario 2:} Providing convenient power to IoT users is a valuable service to help them to stay connected. Wireless energy transfer technologies \cite{na2018energy} transforms the way people charge their IoT devices and enables energy sharing between mobile IoT devices \textit{seamlessly} from a distance. We consider  \textit{crowdsourcing energy as a service} scenario which has the potential to create a \textit{green} environment \cite{lakhdari2018crowdsourcing}. For example, someone wearing smartshoes that have generated energy through walking can share the harvested green energy with other IoT devices wirelessly within a \textit{range}. This can also enable users to recharge their smartphones as they move.

\section{Related Work}

%
%
%
%

We provide an overview of the relevant research in relation to the selection and composition of crowdsourced moving services. We first present a review of \textit{mobile crowdsourcing} frameworks. We then review previous studies on \textit{service composition}. Finally, we survey \textit{co-movement discovery} approaches.

\subsection{Mobile Crowdsourcing}
Crowdsourcing aims at \emph{outsourcing} tasks to the \emph{crowd} for faster and efficient execution. Crowdsourced tasks may require specific requirements compared to others. For example, computing tasks require the crowd to have computers with enough processing power. \emph{Spatial tasks} are tasks that require specific spatial attributes from the crowd for successful task fulfillment. For example, spatial tasks may require the crowd to be physically located at certain locations and collect data. Others may require the crowd to move between different locations to fulfill their task. Crowdsourcing spatial tasks are often referred to as \emph{Mobile Crowdsourcing}. Mobile crowdsourcing typically involves a crowd with mobile devices (e.g., smartphones and smartwatches) to satisfy the spatial requirements by the tasks. There have been several works in the area of mobile crowdsourcing. In this section we focus on two types of mobile crowdsourcing: (1) spatial crowdsourcing \cite{kazemi2012geocrowd, tong2020spatial}, and (2) urban crowdsourcing \cite{shin2012crowdsourcing,marzano2019crowdsourcing}.

In spatial crowdsourcing, the outsourced tasks are typically designated with specific locations. The crowd should execute the tasks at the specified locations. CrowdSensing@Place (CSP) \cite{chon2012automatically} is one example of spatial crowdsourcing. CSP aims at labeling places into categories (e.g., cafe or restaurants). It leverages spatial data, user trajectories, and sampled audio clips and images to achieve this. Another framework is proposed in \cite{bulut2011crowdsourcing} that tries to find a suitable set of users to answer location-based queries. Location-based services (e.g., Foursquare) are used to answer queries instead of relying on spatial task assignments to users. It is worth noting that despite many works proposed on spatial crowdsourcing, there has been little attention to spatio-temporal crowdsourcing.

Urban crowdsourcing aims at detecting users' transportation modes using collected spatio-temporal and acceleration data \cite{shin2012crowdsourcing}. The work in \cite{zimmerman2011field} and \cite{steinfeld2011mobile} proposes mobile crowdsourcing approaches that use commuters trajectories to predict real-time arrival of buses. A new platform, namely OneBusAway, is proposed in \cite{ferris2010onebusaway} to predict the real-time arrival time of buses. OneBusAway collects users' comments and feedback through Twitter, blogs and bug trackers. Another approach is proposed in \cite{said2018mobile} to accurately identify the “right” crowdsourced sensors to answer a particular journey planning request. In particular, an unsupervised learning approach is introduced to select and cluster the right mobile crowdsourced
sensors based on common patterns in their trajectories.

We investigate in our work the concept of \emph{mobile crowdsourced services} that combines mobile crowdsourcing and the service paradigm. We show that mobile crowdsourced services offer more efficient techniques for processing \emph{spatio-temporal sensor data}.

\subsection{Service Composition} 

Mobile crowdsourcing aspires to provide a platform where moving users act as service providers offering crowdsourced services for a smart city. Only a few studies have focused on crowdsourcing as a service \cite{peng2018crowdservice, kantarci2015sensing, murturi2015reference, iosifidis2014enabling}. 
For example, in \cite{peng2018crowdservice} a CrowdService framework is developed to provide crowd worker and crowd intelligence as crowd services via mobile crowdsourcing. A composition approach is developed based on Genetic Algorithm  (GA) to provide near-optimal composite services.
\cite{sheng2012sensing} introduces a  crowdsourced service platform called sensing as a service ($S^2aaS$). In this platform, mobile service providers or smartphone users can request sensing services if they have fulfilled previous sensing tasks \cite{sheng2013sensing}. Furthermore, several crowdsourced service composition frameworks have been proposed to select and compose crowdsourced services. In \cite{peng2016crowdservice}, an agent-based crowd service composition framework is proposed in a scenario of purchasing a secondhand laptop. The framework takes into account users' constraints including the response time and the cost to select and compose crowdsourced services.
A crowdsourced service framework is proposed for composing energy services \cite{lakhdari2020composing}. A new temporal composition algorithm which is a variation of a fractional knapsack algorithm is developed to compose crowdsourced energy services to satisfy a user's energy requirement. However, the studies above assume that the crowdsourced services are fixed in space and time.

 A service composition approach is proposed in \cite{deng2016mobilitye} while considering the service's QoS and mobility. The mobility-aware QoS notion is built using the service invocations' mobility model, which describes the performance of a service. On the other hand, only one QoS criterion i.e., location-sensitive response time is taken into account to select an optimal service composition plan in \cite{dengmobility} and \cite{deng2016mobilitye}.
A Mobile Service Sharing Community (MSSC) is proposed in \cite{dengmobility} for moving service users and providers. Additionally, a composition approach for mobile service is introduced based on Krill-Herd algorithm for finding optimal response time. 
In \cite{wu2019mobility}, a mobility-aware service selection approach is proposed that takes into account a user's movement path. The selection algorithm is modelled based on GA and simulated annealing algorithm. The proposed approach reduces the response time of the service request by selecting appropriate edge servers as the user moves.  
A three-tier IoT service composition framework that takes into account spatio-temporal and energy constraints is proposed in \cite{sun2018energy}. In this framework, an IoT service composition algorithm is developed that adopts GA, Ant Colony Optimization, and Particle Swarm Optimization to find an optimal composition plan while reducing network energy consumption.
In contrast to the above studies, we propose a service composition approach where both service provider and user are moving together.

There are some composition approaches that integrate service selection and compositions with Reinforcement Learning RL \cite{wang2013novel,xu2012multi,gai2018reinforcement}. 
In \cite{jureta2007dynamic}, a novel multicriteria-driven reinforcement learning algorithm is proposed for dynamic Web service composition which adapts Randomized Reinforcement Learning (RRL) \cite{achbany2006optimal}. The proposed approach enables continuous adjustment of the service composition through learning about the quality of new services and exploring new composition plans while optimizing multiple criteria and satisfying users' constraints. 
There are two RL-based service composition algorithms \cite{wang2008preference, moustafa2013multi} which are not efficient for large scale service composition.  
In \cite{wang2014adaptive}, a new model for the large-scale adaptive service composition based on Multi-Agent Reinforcement Learning (MARL) is introduced. The model utilizes the coordination equilibrium and fictitious play process to ensure the convergence of the agent to a unique equilibrium.   
The previous approaches are applied in web service composition. In contrast, our approach is based on spatio-temporal aspects of moving IoT services.

\subsection{Co-Movement Discovery}

There exist many studies to discover similar co-movement patterns including flock \cite{vieira2009line, gudmundsson2006computing}, convoy \cite{jeung2008discovery}, swarm \cite{li2010swarm} and moving clusters \cite{kalnis2005discovering}. A flock pattern refers to a group of at least  $m > 1$ objects moving together within a user-defined  disk with radius $r$ for at least $k > 1$ consecutive timestamps. In the flock pattern discovery, moving objects are clustered based on a disk-based region. A key challenge in flock pattern is the selection of proper disk size. Larger disk size may capture wrong objects and a smaller radius may miss some objects.  To overcome this challenge, a convoy pattern is introduced to cluster moving objects using a density-based clustering method like DBSCAN \cite{ester1996density}. Instead of staying within a disk, moving objects of a convoy are connected based on the density. While flock and convoy patterns have a temporal consecutiveness constraint, swarm and moving cluster patterns relax this constraint through accepting short-term deviations. However, it is not required to have unique (i.e., same) objects throughout the timestamps in moving cluster pattern. 
 Swarm, group \cite{wang2006efficient} and platoon \cite{li2015efficient} adopt different pruning techniques relying on depth-first search which are not efficient in parallel processing \cite{fan2016general}. Most existing methods rely on centralized indexing methods built on top of the whole dataset which may not be effective and efficient in parallel computing \cite{fan2016general}. We focus on the parallel discovery of moving flock patterns without using an indexing method.

\begin{figure}[t]
  \centerline{\psfig{figure=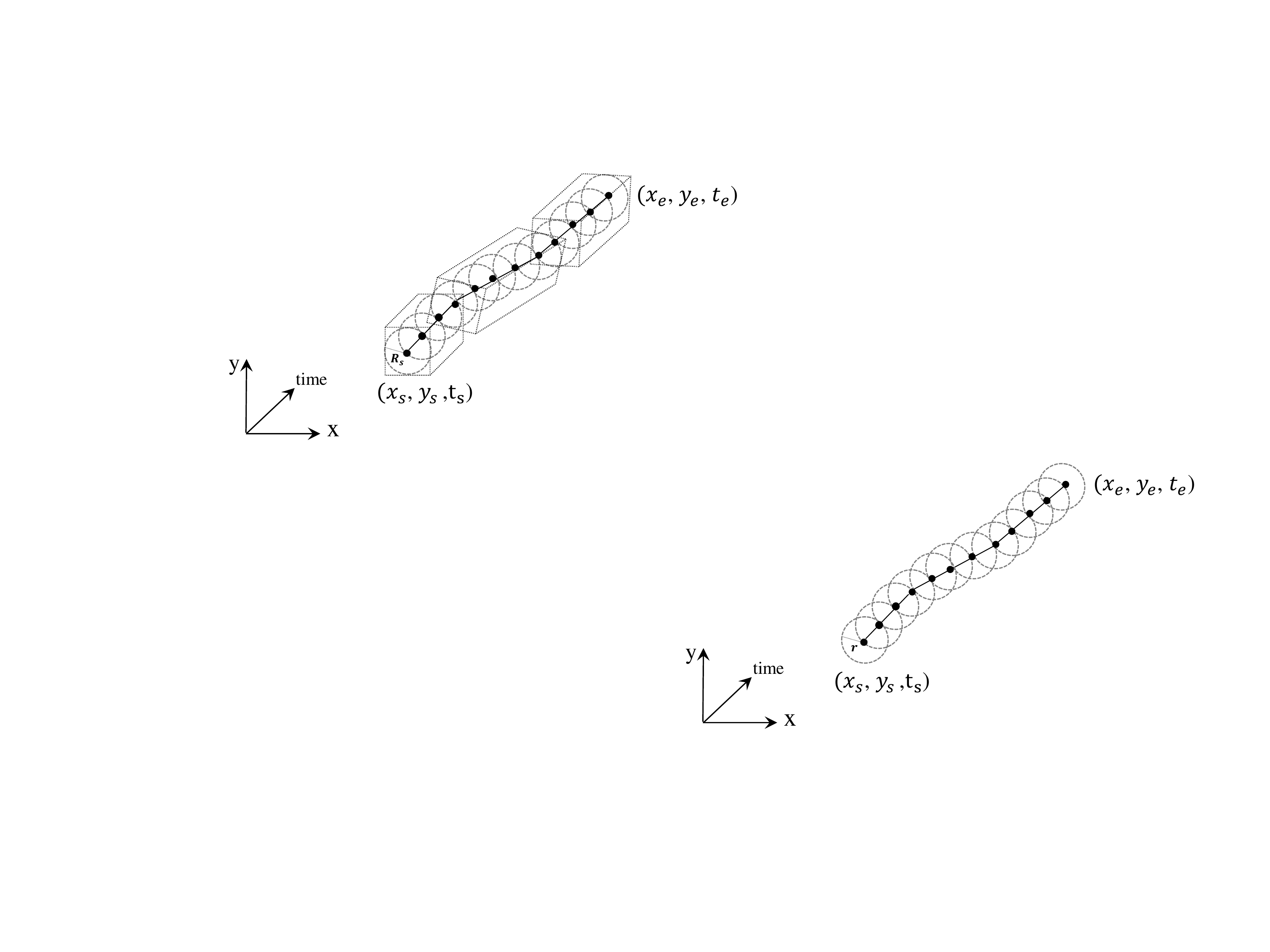,scale=0.40}}
  \caption{ Moving crowdsourced service model.}
  \label{fig:service} 
\end{figure}

\section{System Model and Problem Formulation }\label{sec:Smodel}
In this section, we first formally model crowdsourced moving services by moving regions. We then formulate the problem of selecting and composing the best candidate moving services with respect to their deterministic behavior. 


%

\subsection{Problem Formulation}
\textit{Definition 1:} \textit{ Moving Crowdsourced Service $MS$}. 
   A  moving crowdsourced service $MS$ is a tuple of $<$ $id$, $F$, $Q$ $>$ where 
\begin{itemize}
   \item $id$ is a unique service identifier,
   \item $F$ is a function offered by $MS$, e.g.,  providing a moving WiFi hotspot. The function of the moving service represents a moving service's coverage in space and time. We define the coverage as a moving region which is described by a tuple of $<$ $T_s$, $R_{t_i}(p_i)$ $>$ where 
   \begin{itemize}
  
   \item $T_s$ is a service trajectory which is a sequence of timestamped samples $\{< t_i$, $x_i$, $y_i$ $>$, $1\leqslant i \leqslant k \}$, where $x_i$ is longitude and $y_i$ is latitude and $t_i$ is a timestamp. 
   \item $R_{t_i}(p_i)$ is the specific coverage region that is offered by $MS$. Without loss of generality, the moving region is represented by a circular area which is centered at $p_i$ with the radius $r$ at $t_i$,  
   \end{itemize}
   
   \item $Q$ is a set of QoS attributes $q_i$ ( e.g., capacity). 
   \end{itemize} 
   
Fig. \ref{fig:service} shows a moving crowdsourced service model in a 3D space.



\textit{Definition 2:} \textit{User Trajectory ${T}_u$}. 
A user trajectory is the path that is traveled by a user which is a set of k time-stamped samples ${T}_u $=\{ $<$ $ut_i$, $ux_i$, $uy_i$ $ >$, $1\leqslant i \leqslant k$ $\}$.

%

%
%
%


\textit{Definition 3: Spatial Candidate Pair.} 
Given a set of moving services $ \top  = \{MS_1, MS_2, ... , MS_n\} $, a user trajectory $T_u = \{ up_1 , up_2, ..., up_n \}$ where $up_i$=( $ux_i$, $uy_i$ ), a search radius $r_s$, a moving service is formed as a spatial candidate pair $cp_{t_i}$ for a given timestep $t_i$ if its location $MS_k.p_i$ at $t_i$ is inside a disk region $D_{t_i}(center = up_i(t_i), radius = r_s)$. The service is inside the disk if the Euclidean or Harvestine distance $d(T_u.up_i, MS_k.p_i)$ between two points of user trajectory $T_u.up_i^{t_i}$  and service trajectory $ MS_k.p_i^{t_i}$ at timestep $t_i$ is less than $ r_s $. Without loss of generality, our method can be extended for other distance metrics including network distance and Manhattan distance. The range distance $r$ reflects the maximum spatial proximity allowed.

\textit{Definition 4: Valid Candidate Moving Service.}
A moving service $MS_i$ is a valid candidate service for a given user trajectory if it is paired with the user trajectory over $w$ consecutive timesteps, where $w> 0$, i.e., 
$C_{MS_i}= \{cp_{t_a}, cp_{t_a}, ... cp_{t_w}\}$, $t_a < t_b < ... < t_w$ and $|a-b| =1$. 

\begin{figure}[t!]
  \centerline{\psfig{figure=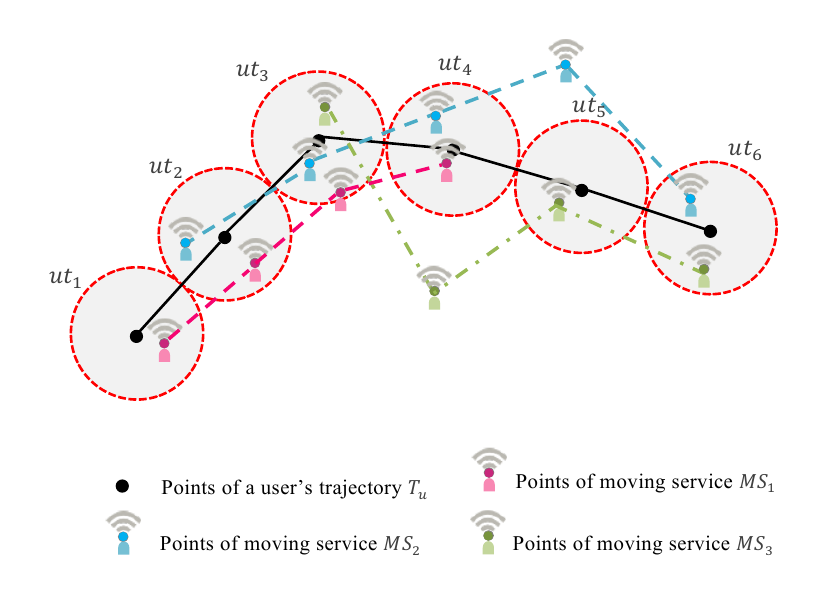,scale=1}}
  \caption{ An example of valid candidate moving services.}
  \label{fig:flocks} 
  \end{figure}

For example,  in Fig. \ref{fig:flocks}, we plot six timestep snapshots of a user moving in an arbitrary trajectory. 
Spatio-temporal neighbour search of a user trajectory at timestep ${ut}_4$ is $MS_1$ and $MS_2$, while $MS_3$ is not a valid candidate moving service. Services in spatial proximity of a user are grouped in circles.

We use the following assumptions in our problem formulation:

\begin{itemize}
    \item One moving service can only serve one user at any point in time.
    \item A moving service moves between any two consecutive timestamps $t_i$ and $t_{i+1}$ with a \textit{constant} speed. As a result, we can determine the position of the moving service at any given time in the time interval [$t_i , t_{i+1}$]. As long as the function of finding the moving service's location is constant time, other speed functions could be considered.
    \item Radii of all coverage regions of services are fixed to a single value. Therefore, we consider the region around each user's trajectory point to find overlapped services. 
    \item  The maximum spatial proximity equals the radius of fixed WiFi hotspot coverage (e.g., 20 m).
    \item We focus on \emph{deterministic} moving crowdsourced services. A deterministic environment can be achieved using several incentive-based approaches (e.g., \cite{zhao2017crowdolr, wang2018min, bliemer2010rewarding, cohen2016using, neiat2019incentive}. For example, our work in \cite{neiat2019incentive} proposes a framework that encourages the movement of crowdsourced IoT service providers from {\em over-supplied} regions to {\em under-supplied} regions.

\end{itemize}

\textbf{Problem Definition.}
Given a set of  moving crowdsourced services  $ \top  = \{MS_1, MS_2, ... , MS_n\} $, a user trajectory ${T}_u$  and a search radius $r_s$ as input, the problem is formulated as finding the ``optimal'' composition plan that gives the best trade-offs among multiple QoS criteria i.e., a high QoS while maintaining a low number of disconnections.
The output is a composition plan $CP$ which is a sequence of moving crowdsourced services $CP = \{S_1, S_2, ... , S_n\}$ $iff$ $S_i$ is a valid candidate moving service for a given user trajectory (see Definition 4).


\subsection{QoS Model for  Moving Crowdsourced Service}\label{sec:qos}

A key challenge is to find a service that offers a better quality due to the diversity of moving services. QoS parameters are used to distinguish among moving services. It is worth noting that the proposed quality model is extensible. A new QoS parameter (either generic or domain-specific) may be added without fundamentally altering the underlying computation mechanisms. For example, in the crowdsourced energy service scenario, we could use energy-related QoS parameters including Transmission Success Rate and Deliverable Energy Capacity which are proposed in \cite{lakhdari2018crowdsourcing} to distinguish among energy services. In our WiFi hotspot scenario, we use one quality attribute that we proposed in \cite{neiat2017crowdsourced} \emph{capacity}.

 \begin{figure}[t]
     \centerline{\psfig{figure=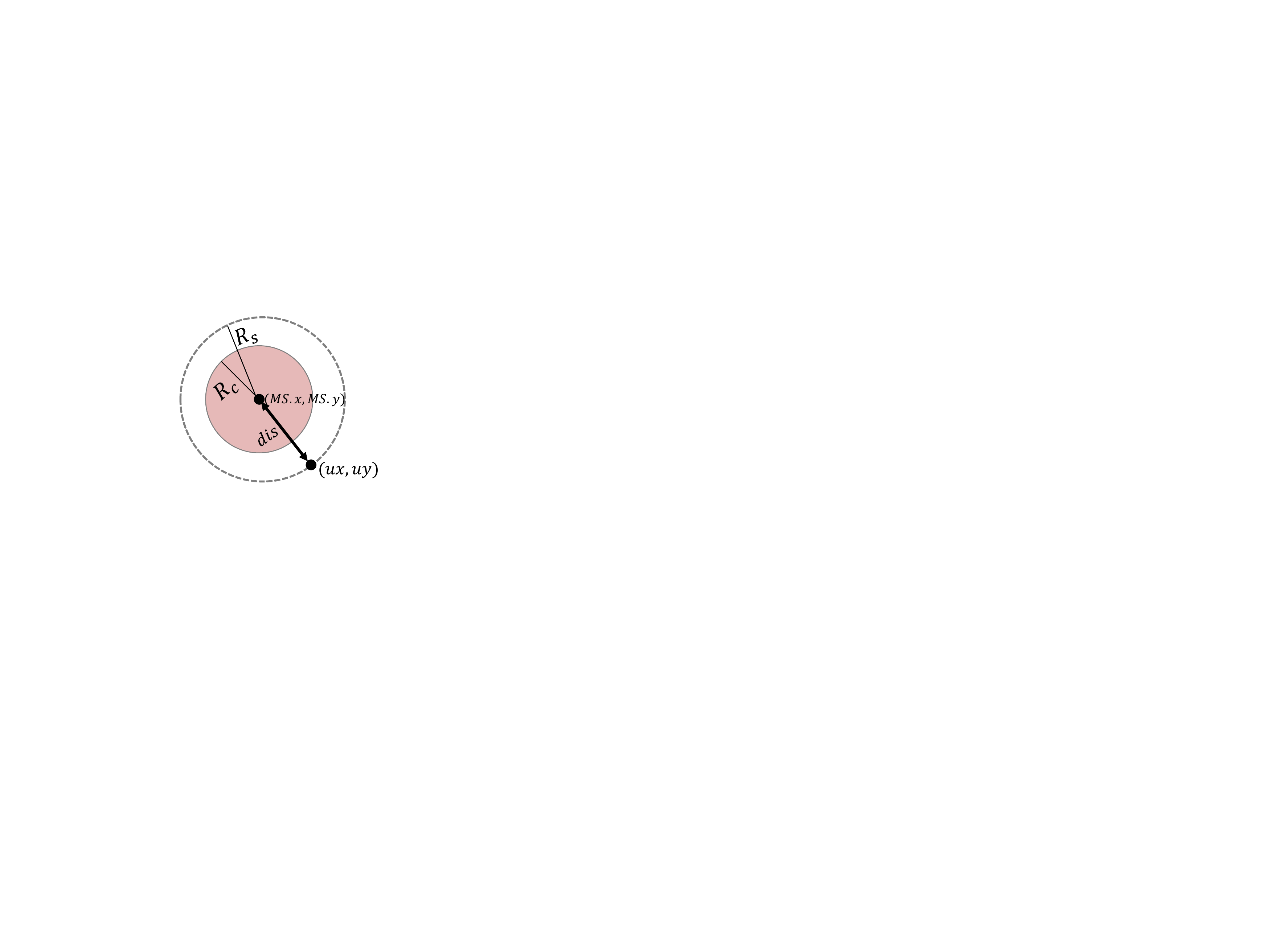,scale=0.44}}
  	 \caption{Strength Model.}
  	\label{fig:qos}
 \end{figure}
 
\emph{Capacity \textit{(cap)}:} Capacity represents the maximum for the information transmission data rate. SNR Shannon-Hartley theorem \cite{shannon2001mathematical} is used to model the capacity. In our model, the capacity is directly proportional to the \textit{signal strength}. The better signal strength is, the higher Signal-to-Noise-Ratio (SNR) i.e., less error is. We assume that the error rate is fixed. Hence, increasing the capacity increases the signal strength, which leads to more successful transmissions. Given a moving crowdsourced service $MS$, $q_{cap}$ is computed as follows.
\begin{equation}
q_{cap}= \frac{B}{K}log_{2}( 1 + str) 
\end{equation}

where $B$ is the total available bandwidth and $K$ is the maximum number of concurrent requests that a moving service $MS$ can support. Total available bandwidth is assumed to be equally allocated between different IoT service users.  
Signal strength $str$ represents the sensing region of a moving crowdsourced service. The strength is computed based on the distance between the user trajectory point $T_u.p_i = (ux,uy)$ and the moving service point $MS.p_i=(MS.x, MS.y)$ at timestep $t_i$. For instance, as the user moves closer to the WiFi hotspot, the perceived WiFi signal gets stronger. The $str$ is based on the exponential attenuation probabilistic coverage model \cite{altinel2008binary}  which is computed as follows.

\begin{equation}
\left\{\begin{matrix}
1 & 0\leqslant pdis(T_u.p_i,MS.p_i)\leqslant R_c\\ 
e^{-kd} & pdis(T_u.p_i,MS.p_i)> R_{c}\\ 
\end{matrix}\right.
\end{equation}

where $d = pdis(T_u.p_i,MS.p_i)- R_c$ and $pdis(T_u.p_i,MS.p_i)$ is the \textit{perpendicular} distance from the crowdsourced service center point $MS.p_i$ to a user trajectory point $T_u.p_i$ (Fig. \ref{fig:qos}). $R_c$ is a confident radius and $k$ is a decay factor which determines the rate of the signal attenuation with regard to the distance. $R_c$  and $k$ parameters can have different values based on the sensor types and the operation environment which can be obtained through experiments. The strength $str$ is within the range of (0,1]. If the user trajectory point is within the distance of $R_c$, the strength is 1 which means full signal. In the interval ($R_s - R_{c}$) where $R_s=pdis(T_u.p_i,MS.p_i)$, the value of $str$ exponentially approaches zero as the perpendicular distance increases. In our model, $str \neq 0$ because a valid candidate crowdsourced service point is paired with a user trajectory point which means that the value of $pdis$ is not beyond $R_{s}$
(Fig. \ref{fig:qos}).

The QoS value of a full composite moving crowdsourced IoT service is computed by calculating the average of the capacities of all its component moving services.

\section{Deep Reinforcement Learning-based Composition Algorithm}

\begin{figure*}[ht]
  \centerline{\psfig{figure=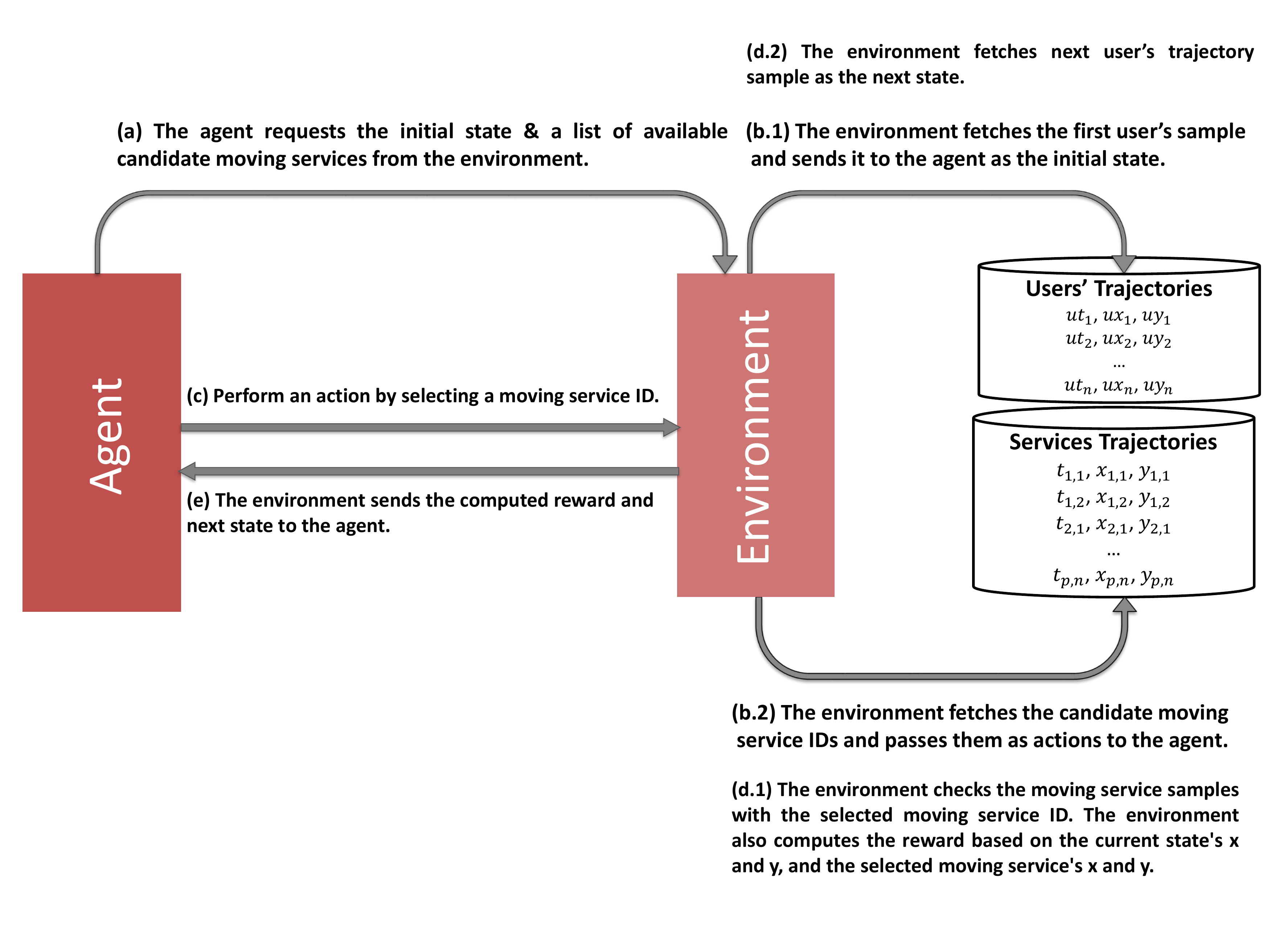,scale=0.38}}
  \caption{ Reinforcement learning for moving crowdsourced service composition.}
  \label{fig:qlearning} 
  \end{figure*}


The proposed framework aims at finding the optimal service composition for a given moving consumer. The framework neither assumes nor requires consumers to change their trajectories for better services. Simply put, the composition plan is obtained by identifying the set of services that \emph{intersects} with the consumer's trajectory. Performing the selection in real-time may disrupt the user experience for consumers. For example, if we assume a consumer that wishes to use a WiFi hotspot for watching videos. The playback may be interrupted every now and then whenever the device is looking for a new service. Therefore, we opt for ahead of time service selection to minimize the interruption overhead.

One single moving service may not fully cover a user trajectory as the user may not necessarily share their entire route with said moving service. As a result, we need to compose candidate moving services to satisfy a user's requirements and ensure service continuity. In this section, we present our approach for selecting and composing all candidate moving services. 
We leverage \emph{Deep Reinforcement Learning DRL}  to find and compose valid candidate services overlapping with the user trajectory. Reinforcement Learning is a subclass of machine learning, where an \emph{agent} learns about an \emph{environment's} behavior through \emph{explorations}. The main reason for our choice to DRL is its ability to discover the "cumulative" optimal set of service trajectories, given the trajectory segments of a user. The DRL does that by assigning rewards for every action the agent invokes. In our work, the actions are the service segments that a user can use. The agent's role is to pick the next service segment that would maximize the overall reward. Therefore, the agent should not only consider the current service and user trajectory segments to make the selection but also future service and user segments. DRL offers a way for the agent to predict the "future" (i.e., unknown knowledge) by learning the behavior of the environment through explorations. 
The agent is the \textit{composition} algorithm which determines the optimal combination of candidate moving services over time. An environment is characterized by its set of \emph{states} ($S$) and \emph{actions} ($A$). In our work, the environment is the combination of service trajectories and a single user trajectory, (see Fig. \ref{fig:qlearning}). A user trajectory consists of a series of samples i.e. states. Each sample is a tuple: $<ut, ux, uy>$, where $ut$ is the timestamp, $ux$ is the longitude coordinate, and $uy$ is the latitude coordinate. The services trajectories dataset comprises a series of records representing different samples $<t, x, y>$ for different services. A sample in the services trajectories dataset contains the same information as a user trajectory sample with the addition of a service ID and QoS attributes (see Definition 1). 
At any given time, the environment has a current state $s \in S$ i.e. the current user trajectory sample that is reported to the agent. The agent selects an \emph{action} ($a \in A$) i.e., a valid candidate moving IoT service to invoke on the environment. Upon action invocation, the environment generates a \emph{reward} ($r$) based on the action and its current state. Additionally, the environment changes its state according to the invoked action and previous state. The reward and next state are reported back to the agent i.e., composer.

More formally, we define a composite moving service as follows.

\textit{Definition 5:}  \textit{ Composite Moving Crowdsourced Service.}
A composite moving service {CS} is a sequence of component moving services, which is defined as a 5-tuple of $<S^g,s_0^g,s_r^g,A^g(s),R^g>$ where

\begin{itemize}
    \item $S^g$ is a finite set of states i.e., samples $<t, x, y>$ that are observed by an agent g;
    \item $s_0^g \in S^g$ is the initial state of the agent $g$ and the execution of the composite moving service starts from this state. Here the initial state is the first user trajectory sample. 
    \item $s_r^g \subset S^g$ is the set of terminal states where the execution of a composite service terminates upon arriving at one of the states. The terminal state is the last sample of a user trajectory. 
    \item $A^g(s)$ is the set of actions that are taken at each state $s$. Here we replace the actions with valid candidate moving services (see Definition 4). At each state, we have a set of moving services that could be selected and executed. 
    \item $R^g$ is the reward function when a moving service is invoked. There are multiple QoS objectives which the agent wants to achieve. The user receives the reward i.e., desired QoS when the agent moves to the next state $s$ from $s'$. Therefore, the reward is computed using the QoS of a composite service. 


\end{itemize}

The composition algorithm aims at finding the \emph{optimal policy} ($\pi^*$) which is defined as the procedure for selecting candidate moving services (i.e., the action to be invoked by the agent in each state). This guides the agent toward an optimal set of moving services that gives the best trade-offs among multiple QoS criteria i.e., maximum accumulated reward. 
More formally, the optimal policy is expressed as follows:
\begin{equation}
    \pi^*(s) = a
\end{equation}
where $s \in S$ is the environment's state and $a \in A$ is an optimal moving service (i.e., action) to invoke. Obtaining the optimal policy is achieved by solving the \emph{Q-value function}:
\begin{equation}
    Q(s, a) = r(s, a) + \gamma \sum\limits_{s'} \max\limits_{a'} Q_{i}(s', q') | s,a
    \label{eq:q_value}
\end{equation}
where $Q(s, a)$ is the accumulated QoS reward, given an agent starting at state $s$ and invoking action $a$, $r(s, a)$ is the instantaneous reward when invoking action $a$ at state $s$, and $\gamma$ is a discounting factor.

Finding the optimal policy directly using Equation \ref{eq:q_value} is impractical due to a potentially large number of states and actions. As a result, the number of state-action combinations increases drastically which in turn leads to exponential time complexity. Therefore, Neural Networks are used in conjunction with Equation \ref{eq:q_value} to build a policy model. The use of Neural Networks to find the optimal policy is referred to it as \emph{Q-Learning}. The inputs to the Neural Network are the different parameters defining a particular state $s$, and its outputs are all possible actions that the agent can take.

\begin{algorithm}
	\fontsize{10pt}{10pt}
	\caption{  Reinforcement Learning-based Composition Training Algorithm}
	\textbf{Input:} A set of sampled moving service trajectories $T_s$, a set of sampled user trajectories $T_u$, and the radius $r$\\
	\textbf{Output:} A model for policy $\pi$
	\label{alg:rl}
	\begin{algorithmic}[1]
		\State // Initialization
		\State $\epsilon$ $\leftarrow$ 1.0
		\State $model$ $\leftarrow$ create an initial neural network model
		\State $memory$ $\leftarrow$ []
		\State $env$ $\leftarrow$ initialize environment using $T_s$
		\State // Model Training
		\For {$t_u \in T_u$}
			\For { $i$ $\leftarrow$ 1 \textbf{to} repetition }
				\State $env$.$current\_user$ = $t_u$
				\State $state$ $\leftarrow$ $t_u$
				\If{random() $<$ $\epsilon$}
					\State $action$ $\leftarrow$ pick a random service id from $T_s$
				\Else
					\State $action$ $\leftarrow$ predict the next action using $model$ and $state$
				\EndIf
				\State $reward$, $next\_state$ $\leftarrow$ act on $env$ using $action$
				\State store $next\_state$ and $reward$ in $memory$
			\EndFor
			\If{$memory$ is full}
				\State train $model$ using data in $memory$
				\State $\epsilon = \epsilon * 0.995$ // gradually decreasing exploration
			\EndIf
		\EndFor

		\State \textbf{return} $model$
	\end{algorithmic}
\end{algorithm}

  \begin{figure*}[ht]
  \centerline{\psfig{figure= 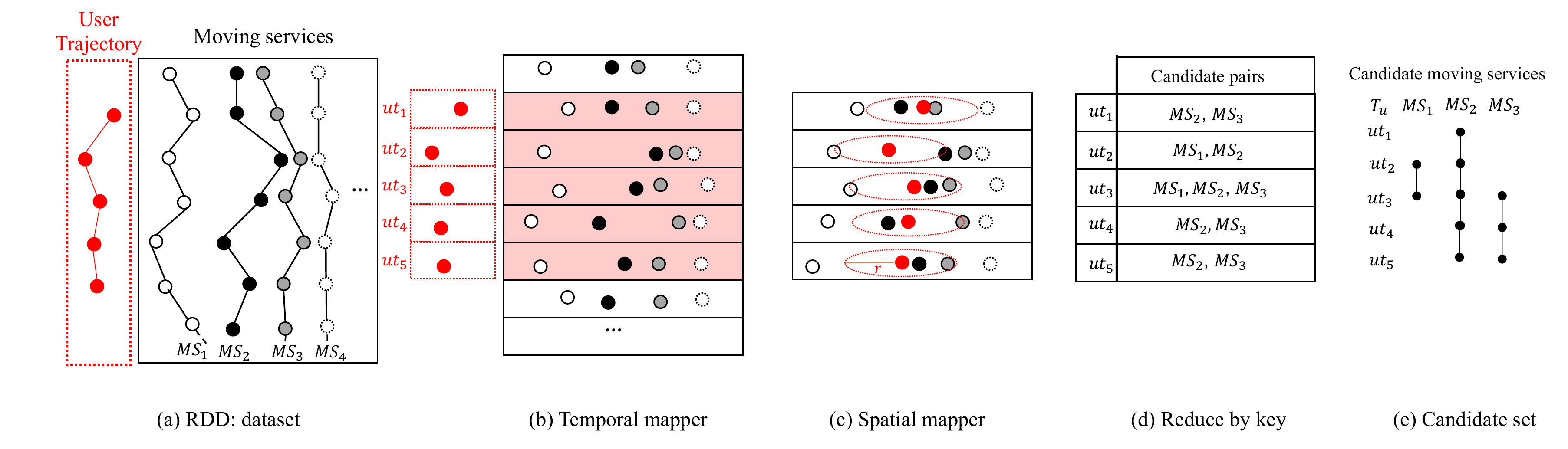, scale=0.50}}
  \caption{ Workflow of parallel flock-based moving crowdsourced  service discovery.}
  \label{fig:mapreduce} 
  \vspace{-0.4cm}
  \end{figure*}

Q-learning has two main phases: \emph{exploration} and \emph{exploitation}. In the exploration phase, the agent selects random actions and keeps records of the resulting rewards and next states. The collected records are then fed into a Neural Network to train the policy. During the exploitation phase, the agent uses the trained model to predict the next action to take. It is worth noting that in the exploitation phase, the agent keeps exploring to adjust the trained model for higher accuracies. 

Fig. \ref{fig:qlearning} summarizes the process. An agent starts by requesting the initial state of the environment (step (a)). The environment fetches the first user trajectory sample and reports it as its current state (step (b.1)). Additionally, the environment fetches the list of service IDs and reports them to the agent as possible actions to invoke (step (b.2)). During exploration, the agent randomly invokes an action (by selecting a valid candidate moving service ID) on the environment (step (c)). The environment computes the reward based on the invoked action. We set the reward as the QoS provided by the selected candidate moving service (step (d.1)).
The environment updates its current state by setting it to the next user trajectory sample (step (d.2)). The next state and reward are sent to the agent (step (e)). The agent continues invoking actions until all samples in the user trajectories are visited. Upon traversing all samples of a user trajectory, the environment resets itself (by setting its state to the first user trajectory sample), and the process repeats. Note that several user trajectories should be used to increase the prediction accuracy of the generated model.

\textit{Two cases have not been addressed in our previous scenario: (1) no valid candidate moving services for a given user trajectory sample, and (2) an agent selects a moving service that is not a valid candidate (i.e., does not overlap either in time or space with the current user trajectory sample)}. To resolve the first case, we introduce the concept of a \emph{dummy service}. A dummy service essentially means that there are no valid candidate moving services for a particular user trajectory sample. The goal of a dummy service is to divert the agent from selecting an invalid candidate moving service when no overlapping services are found. Accomplishing this is carried out by giving a lower reward value whenever the dummy service is selected, e.g., -1 as opposed to [0 - 1] for normal reward values. Resolving the second case is achieved by penalizing the agent when an invalid moving service is selected, e.g., -10. By this, an agent will always favor selecting the dummy service over an invalid one since it has a higher reward value.

Algorithm \ref{alg:rl} summarizes the training phase of the approach. The algorithm starts by initializing the parameters (Lines 1-5). Initially, the reinforcement learning agent performs exploration only. The exploration to exploitation ratio is controlled by changing $\epsilon$ (1.0 for full exploration and 0.0 for full exploitation). A neural network model is created with random weights in Line 3. The algorithm also initializes an empty space that represents the agent's \emph{memory} (Line 4). The memory stores information that the agent collects during exploration. More specifically, the memory holds every action the agent takes on the environment. Additionally, it stores the rewards the environment assigns for the taken actions. Finally, the environment is set up using the provided service trajectories $T_s$ (Line 5). The environment uses the service trajectories $T_s$ to determine its set of possible actions as well as the reward for each taken action. 

The model is trained gradually while the agent is exploring the environment and exploiting its current model (Lines 7 through 21). The agent at the beginning favors exploring the environment over exploiting its model. The algorithm loops through each user trajectory in $T_s$ (Line 7). The state of the environment is set using the samples in a given user trajectory. When the agent takes an action, the environment sets its next state to the next sample in the current user trajectory. We allow the environment to exploit each user trajectory $repetition$ times (Lines 8 - 10). In other words, a state can be repeated several times. This allows the agent to experiment with different actions given the same state, and observe the rewards associated with each state-action combination. The agent invokes an action on the environment (Lines 11 - 15). The action is the service to choose given a user trajectory sample. The taken action is based on the value of $\epsilon$. The agent tends to take random actions when $\epsilon$ has a high value (exploration). Conversely, the agent uses its trained model to decide which action to take. Upon each invoked action, the environment changes its state and returns a reward to the agent (Line 16). The reward is generated based on the QoS of the selected service (i.e., the taken action). The reward and next state values are stored in $memory$ (Line 17). The collected actions, states, and rewards in $memory$ are used to train the model (Line 20). In other words, $memory$ guides the training process towards building a model that favors actions with higher rewards. The value of $\epsilon$ is decayed after each training process (Line 21). Decaying $\epsilon$ makes the agent use the trained model more, essentially leading the agent to invoke actions that may have higher rewards.

It is worth mentioning that model training and storage is carried out using \emph{edge servers}. We assume edge servers are conveniently set up to be accessed by moving IoT services. Each edge server is responsible for serving a small subset of moving devices. Therefore, storage and processing overheads are negligible. 

\section{Ground-Truth: Parallel Flock-Based Moving Service Discovery}
We propose a \emph{brute-force} approach to find the optimal composition of moving services for a given consumer trajectory. Our approach is used in our experiments to evaluate the accuracy of our deep reinforcement learning-based composition algorithm proposed earlier. In other words, the brute-force approach acts as a \emph{ground-truth} (baseline) to validate the results obtained using the deep reinforcement learning-based algorithm.

The brute-force is the optimal solution as it performs exhaustive search to find all possible co-moving services with a given consumer's trajectory. We achieve this by identifying \emph{co-movement patterns}. Several studies have been proposed to model moving objects including flock \cite{vieira2009line, gudmundsson2006computing}, convoy \cite{jeung2008discovery}, swarm \cite{li2010swarm} and travelling companion \cite{tang2012discovery}.  All of these group movement patterns require the group to contain the same set of individuals during its lifetime \cite{naserian2018framework}. We opt for \textit{the flock pattern} as it is the most appropriate group movement pattern to describe our WiFi hotspot sharing scenario due to the fixed size of the radius (i.e., WiFi hotspot range). However, other co-movement patterns may be more appropriate for other types of applications.

\begin{algorithm}
   \fontsize{10pt}{10pt}
  \caption{  Parallel Flock-Based Service Discovery Algorithm }
  \textbf{Input:} A set of moving services $\varsigma$ , A user travel trajectory $T_u$, Radius $r$\\
  \textbf{Output:} A set of spatial candidate pairs 
  \label{alg:mapreduce}
  \begin{algorithmic}[1] 
  \State C $\leftarrow$ \{\}
  \State compute a list of  $<t_i$,$ST_{t_i}>$ pairs of $\varsigma$
  \State compute a list of timesteps $<ut_i$,$T_{up_i}>$ pairs of $T_u$
  \State  -- Temporal Mapper Phase--
   \For {\textbf{all} key-value pairs $<t_i$,$ST_{t_i}>$}      
      \State left outer join based on timesteps $ut_i$ 
   \EndFor
  \State -- Spatial Mapper Phase --
  \For {\textbf{all} $t_i$ in key-value pairs $<t_i$,$ST_{{p_i}}>$}      
      \State find spatial candidate pairs based on $r$ and $T_{up_i}$
  \EndFor
  \State -- Reduce Phase--
  \State group-by $ut_i$  
  \State \textbf{Return} spatial candidate pairs set
  \end{algorithmic}
\end{algorithm}
We deploy a parallel moving service discovery approach. We use Apache Spark as a platform for parallel discovery. The Apache Spark employs MapReduce to handle the scalability and fault tolerance issues. Fig. \ref{fig:mapreduce} shows MapReduce jobs which are performed in a sequential workflow. First, the \textit{temporal map phase} conducts temporal pruning of moving services with regards to a user trajectory. In this temporal mapper, timestamp and sampled location are respectively treated as \textit{key} and \textit{value} for each moving service trajectories. Since a user trajectory is important to find the co-movement service, we filter moving service trajectories based on a user trajectory's timesteps. In this regard, we use left outer join as a temporal filtering step to select all moving services that include the user trajectory timesteps (Lines 5-7 Algorithm \ref{alg:mapreduce}). 
Second, for each timestep of a user trajectory, the \textit{spatial map phase} is performed over all filtered sub-trajectories of moving services. In the spatial mapper, we retrieve all spatial candidate service pairs (see Definition 2) whose points are inside a range distance $r$ of the user location point in each timestep. Finally, the spatial mapper outputs candidate pairs in each timestep of the user trajectory. It is represented by a list of key-value pairs $<ut_i$, $ST_i>$, where $ut_i$ is the user's timestep and $ST_i$ is a set of candidate moving service IDs that are paired at $ut_i$ (Lines 9-11 Algorithm \ref{alg:mapreduce}). For each discovered candidate pairs, QoS values are computed. We then extend the discovered candidate pairs to discovered candidate moving services. All candidate moving services are validated based on Definition 3 and invalid candidates are disregarded. 


  \begin{figure}[t]
  \centerline{\psfig{figure=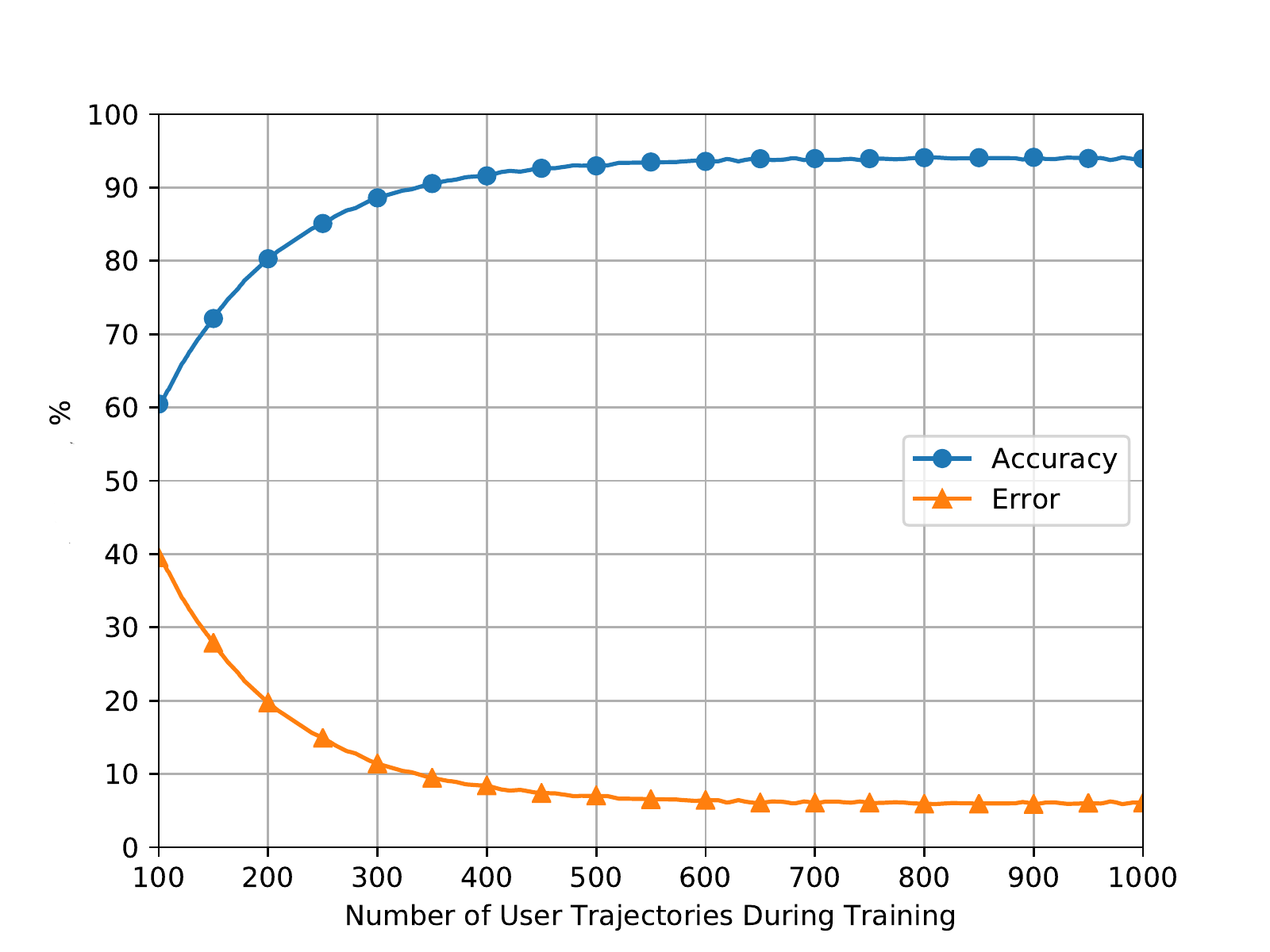,scale=0.48}}
  \caption{ Accuracy on the indoor dataset}
  \label{fig:accindoor} 
  \vspace{ 0.3 cm}
\end{figure}

\begin{figure}[t!]
  \centerline{\psfig{figure=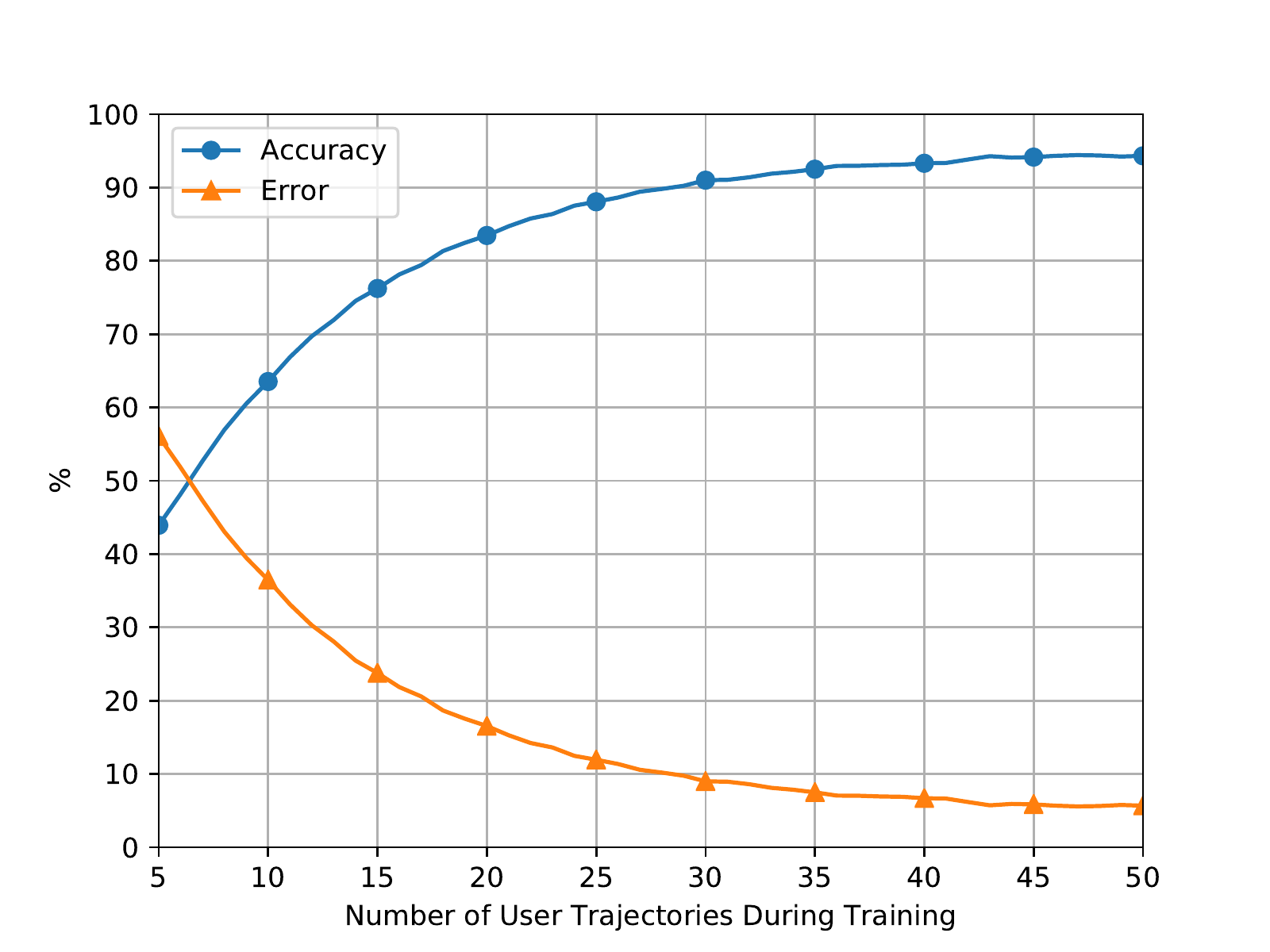,scale=0.48}}
  \caption{ Accuracy on Illinois dataset.}
  \label{fig:accillinois} 
 \vspace{ 0.3 cm}
\end{figure}



\section{Experimental Evaluations}

We evaluate the accuracy and efficiency of our proposed deep reinforcement learning-based composition algorithm. We utilize our ground-truth approach discussed earlier to validate the results and assess the accuracy of our algorithm. We leverage real pedestrian trajectory datasets throughout our experiments. 

\subsection{Experiment Setup}

All experiments are conducted in a cluster with six nodes on Amazon Web Services. We pick one master node and five slave nodes. The master node has a dual-core processor with 4GB memory. The slave nodes are identical, each equipped with a 16-core processor and 64GB memory. We configure the cluster into 15 Spark executors, each taking 19GB memory and 5 cores. 

We use a dense fully connected network to train our model. The inputs to the network are the parameters defining a state whereas the outputs are the actions, which the agent can take. Three hidden layers are used with 512 neurons each. The rectified linear unit (ReLU) activation function is used in all hidden layers. We use dropout with probability 0.5 on all hidden layers to reduce overfitting. The Q-learning's discount factor $\gamma$ is set to 0.9, whereas the Neural Network's learning rate is 0.001.

The reward function used by the deep learning-based composition algorithm is based on the capacity QoS parameter. We assume that the Q-learning algorithm does not have prior knowledge about QoS attributes of moving services since they are computed based on the distance between a user and a moving service. As a result, we rely on the Q-Learning algorithm to learn the optimal execution policy. We use 70\% of the data in each dataset for the training set and the remaining 30\%  for the test set.

We use two real \emph{pedestrian} trajectory datasets. A single trajectory in the datasets is represented with a series of location samples. Each sample represents the location of a person at a specific time. We use the trajectories in the two datasets to represent WiFi hotspot moving services and user trajectories\footnote{To the best of our knowledge, there are no publicly available crowdsourced hotspot/energy environment datasets.}. The trajectories in the datasets are split into two groups. The first group represents the available WiFi hotspot moving services. The second group is considered as the trajectories of the users. Throughout our experiments, we use our proposed algorithm to perform service selection and composition using the moving services and user trajectories groups. In other words, for a given user trajectory, our proposed algorithm aims at finding a subset of moving services that intersect with the user trajectory while maximizes the QoS. The two datasets are described as follows:
\begin{itemize}
    \item \textit{Indoor \footnote{https://irc.atr.jp/crest2010\_HRI/ATC\_dataset}:} The dataset keeps visitors' trajectories in the ATC shopping center in Osaka. The visitors' locations are sampled every 0.03 - 0.06 seconds. In this dataset, we replace sampled timestamps with global sequences that start from 1 to find co-movement patterns. A fixed sampling rate of 0.04 seconds is set, since trajectories do not have synchronized sampled time. We adopt linear interpolation to fill missing points. The dataset contains 1,777,297,164 samples and 185,554 trajectories i.e. moving services. 
    
    \item \textit{Illinois \footnote{https://www.cs.uic.edu/~boxu/mp2p/gps\_data.html}:} The dataset holds six months trajectories from the daily commute of two members in Argonne National Laboratory of the University of Illinois at Chicago. Each trajectory shows a continuous daily trip of a member in Cook County and/or the Dupage County of Illinois. We treat each trajectory as a moving service. The member's locations are strictly sampled every second. There are 357,706 samples and 207 trajectories i.e. moving services. 
\end{itemize}



\begin{figure}[t]
  \centerline{\psfig{figure=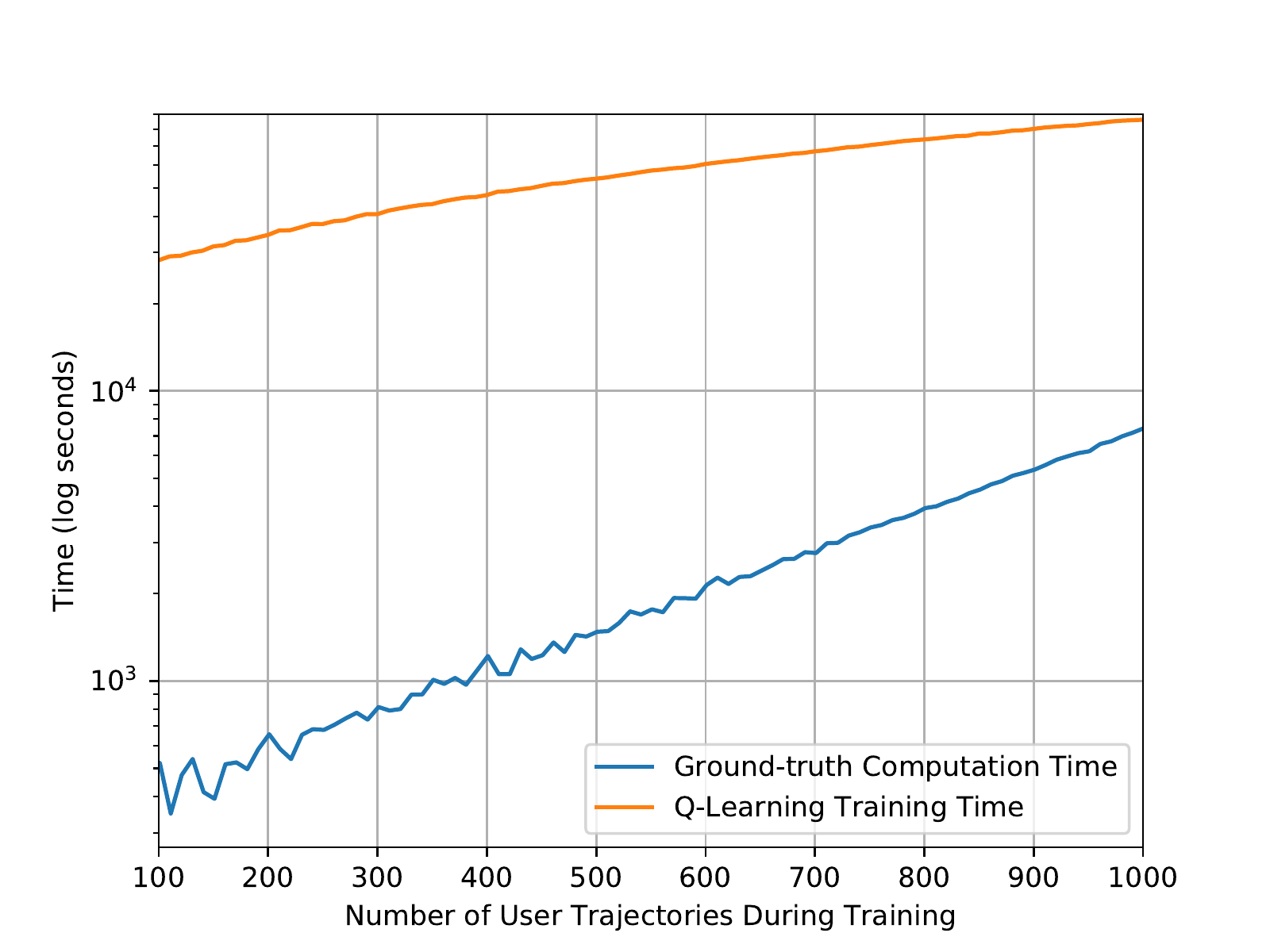,scale=0.46}}
  \caption{Computation time vs. No. of user trajectories.}
  \label{fig:scal2} 
 \vspace{ 0.3 cm}
\end{figure}

To the best of our knowledge, there is limited research investigating QoS-aware moving IoT service composition. We compare the proposed deep reinforcement learning-based composition algorithm to the proposed ground-truth approach to show the accuracy of the selection process.

\subsection{Experiments results}
In this section, we evaluate our deep reinforcement learning-based composition algorithm from three aspects: (1) accuracy; (2) scalability and (3) efficiency of learning.

\subsubsection*{Accuracy}
In the first set of experiments, we study the accuracy of the deep reinforcement
learning-based composition algorithm by comparing it to the ground-truth approach on two real datasets. We evaluate the accuracy of the proposed approach by comparing the deep reinforcement learning-based composition algorithm in selecting valid candidate moving services in each timestep and
their corresponding valid candidate moving services that are retrieved by the ground-truth. 
The accuracy is the ratio between the number of times an optimal service was selected $cs$ to the total number of available valid samples $ns$.

\begin{equation}
    \label{eq:accuracy}
    Accuracy =\frac{|cs|}{|ns|}
\end{equation}

Fig. \ref{fig:accindoor} shows the accuracy (blue curve) and error (orange curve) of the deep reinforcement
learning-based composition algorithm on indoor dataset while varying the number of user trajectories from 100 to 1,000. As mentioned earlier, the accuracy (and error) results are obtained by validating the proposed approach results with the ground-truth approach, which generates the best possible composition plan (i.e., accuracy 100\%). The results show the proposed approach scores high accuracy. As can be seen, the accuracy significantly increases until it reaches around 95\% by 500 user trajectories and then it remains stable. As expected, the accuracy is lower when the number of trajectories is low. The reason is that the lower the samples are, the less accurate the result is. 
Similarly, Fig. \ref{fig:accillinois} shows that the accuracy (blue curve) and error (orange curve) of the composition algorithm on the Illinois dataset. The accuracy increases until it reaches a high accuracy of around 93\%  after 35 user trajectories.

\begin{figure}[t!]
  \centerline{\psfig{figure=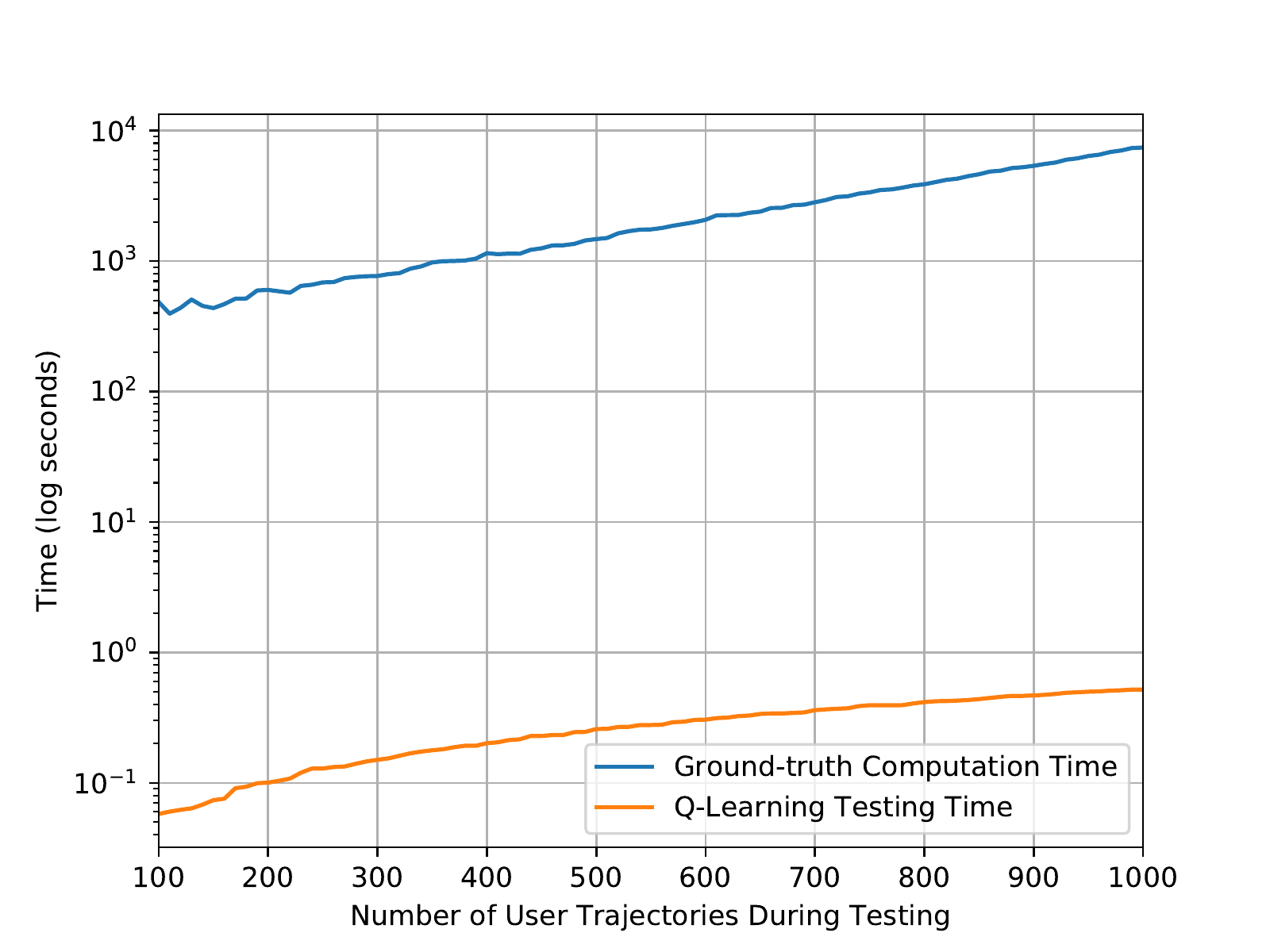,scale=0.47}}
  \caption{Computation time vs. No. of user trajectories.}
  \label{fig:scal} 
  \vspace{ 0.5 cm}
\end{figure}

\begin{figure}[t]
  \centerline{\psfig{figure=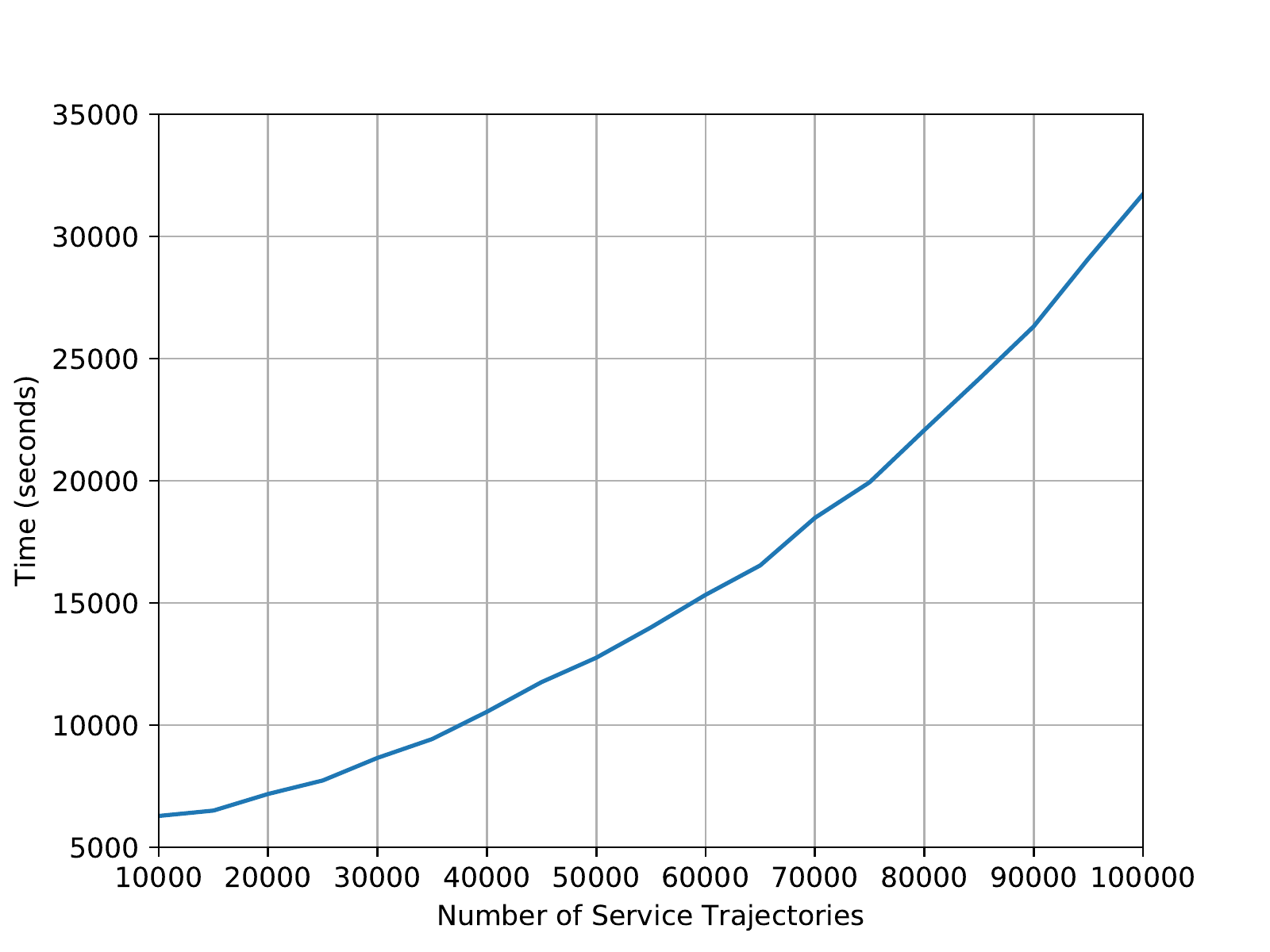,scale=0.45}}
  \caption{ Convergence time vs. No. of moving services.}
  \label{fig:convindoor} 
  \vspace{ 0.5 cm}
\end{figure}

\subsubsection*{Scalability}
In the second set of experiments, we evaluate the scalability of the proposed approach in terms of training and selection computation time. Fig. \ref{fig:scal2} demonstrates the ground-truth computation time in comparison with the Q-learning model training time. As expected, the results show that the ground-truth computation time is significantly lower than the training time. On the other hand, Fig. \ref{fig:scal} illustrates the computation time to find valid candidate services of deep reinforcement learning-based composition algorithm significantly outperforms the ground-truth (i.e. less than 0.1 Sec in Q-learning in comparison with close to 10,000 Sec in ground-truth for 1000 services). This indicates that the deep reinforcement learning-based composition algorithm can select valid candidates much faster than the spatio-temporal MapReduce approach.

\subsubsection*{Efficiency of learning}
\begin{figure}[t!]
  \centerline{\psfig{figure=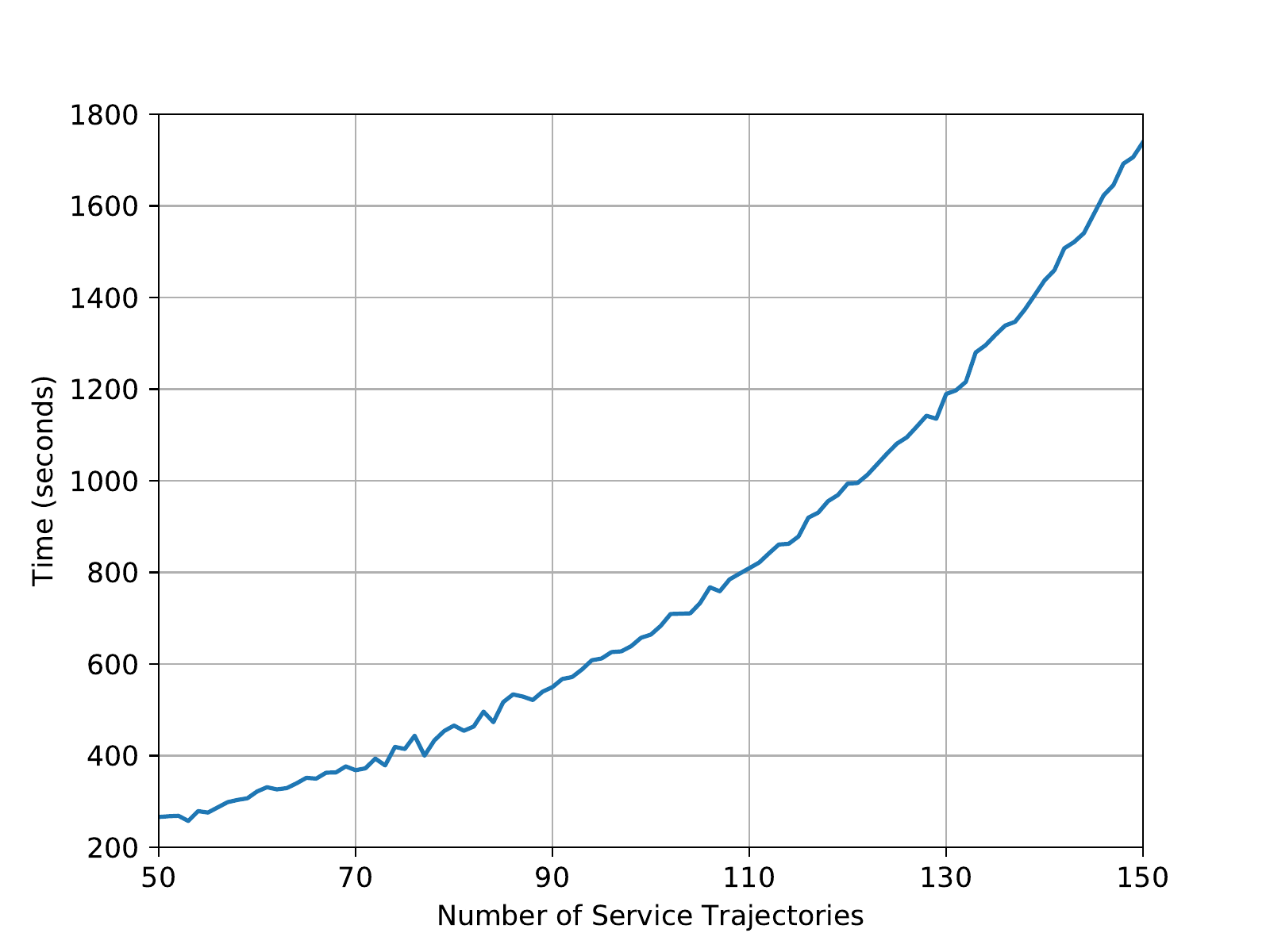,scale=0.46}}
  \caption{ Convergence time vs. No. of moving services.}
  \label{fig:convillinois} 
  \vspace{ 0.5 cm}
\end{figure}

In the third set of experiments, we study the learning speed with increasing the number of moving services on indoor and Illinois datasets. Firstly, we vary the number of moving services from 10,000 to 100,000 on the indoor dataset. The results in  Fig. \ref{fig:convindoor} illustrate how fast the algorithm converges to the optimal composition plan during the learning phase. The convergence time increases polynomially with the increasing number of moving services which is an expected result. This is because the number of candidate moving services that should be searched at each state increases exponentially. Additionally, the convergence time increases slower than the number of moving services. Secondly, we vary the number of moving services from 50 to 150 on the Illinois dataset. Fig. \ref{fig:convillinois} shows that the convergence time increases with increasing the number of moving services. The results also show that our model trains relatively fast to find optimal composition plans (e.g., less than 30 min for 150 services).

\section{Conclusion}\label{sec:conclusion}

We proposed a crowdsourced IoT service framework to select and compose moving crowdsourced IoT services based on spatio-temporal factors. We developed a deep reinforcement learning-based algorithm to select and compose moving services considering QoS parameters without using an index. We also developed a spatio-temporal MapReduce based on flock patterns using Apache Spark to discover moving services as a ground-truth. Our experiments show the scalability and high accuracy of the proposed approach in comparison with the ground-truth. In our future work, we develop and test our proposed approach on different motion patterns, i.e., transportation modes. We also plan to extend the proposed framework to temporal non-deterministic moving services.

\section*{Acknowledgments}
This research was partly made possible by DP160103595 and
LE180100158 grants from the Australian Research Council.
The statements made herein are solely the responsibility of
the authors. The authors also acknowledge the University of Sydney HPC service at The University of Sydney for providing HPC resources that have contributed to the research results reported within this paper.


\bibliographystyle{IEEEtran}
\bibliography{mybib2}

\begin{IEEEbiography}[{\includegraphics[width=1in,height=1.25in,clip,keepaspectratio]{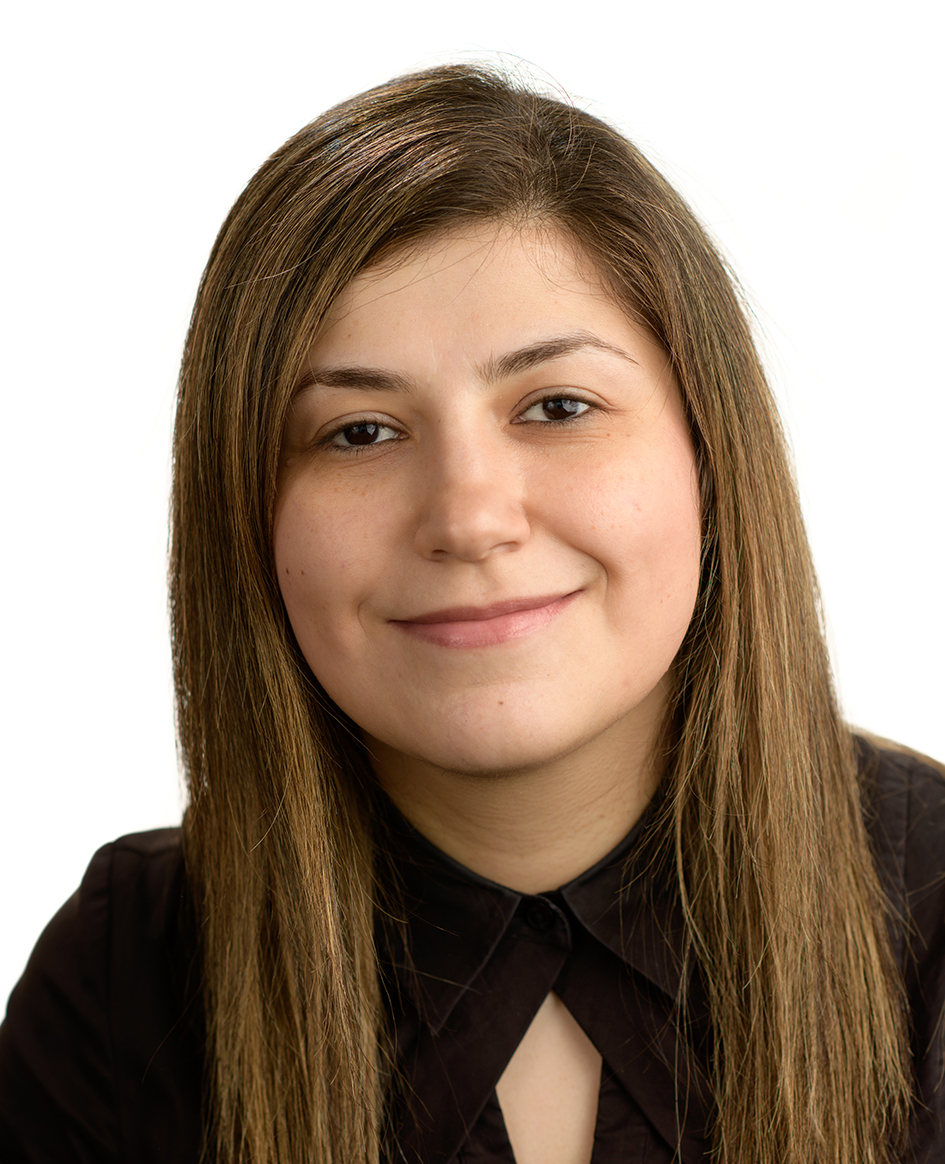}}]{Azadeh Ghari Neiat}
is a lecturer in the School of Information Technology at the Deakin University.  Before joining Deakin University, she was a postdoctoral research fellow at the University of Sydney since 2017. She was awarded a PhD in computer science at RMIT University, Australia in 2017. She has published in top journals and conferences such as CACM, IEEE TKDE, IEEE TSC, ACM TOIT, Future Generation Computer Systems, ICSOC, ICWS, MobiQuitous etc. Her research interest lies at the intersections of Mobile Crowdsourcing, AI, IoT, and Spatio-Temporal Data Analysis.
\end{IEEEbiography}

\vspace{-3cm}
\begin{IEEEbiography}[{\includegraphics[width=1in,height=1.25in,clip,keepaspectratio]{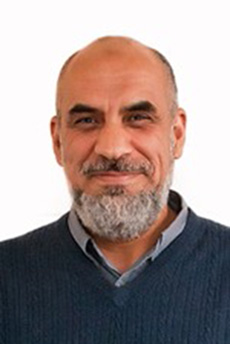}}]{Athman Bouguettaya}
is Professor of Computer Science at the University of Sydney,
Sydney, Australia.He received his PhD in Computer Science from the University of Colorado at Boulder (USA) in 1992. He is or has been on the editorial boards of several journals including, the IEEE Transactions on Services Computing, ACM Transactions on Internet Technology, the International Journal on Next Generation Computing, VLDB Journal, Distributed and Parallel Databases Journal, and the International Journal of Cooperative Information Systems. 
He has published more than 200 books, book chapters, and articles in journals and conferences in the area of databases and service computing. He is a Fellow of the IEEE and a Distinguished Scientist of the ACM.

\end{IEEEbiography}

\vspace{-3cm}
\begin{IEEEbiography}[{\includegraphics[width=1in,height=1.25in,clip,keepaspectratio]{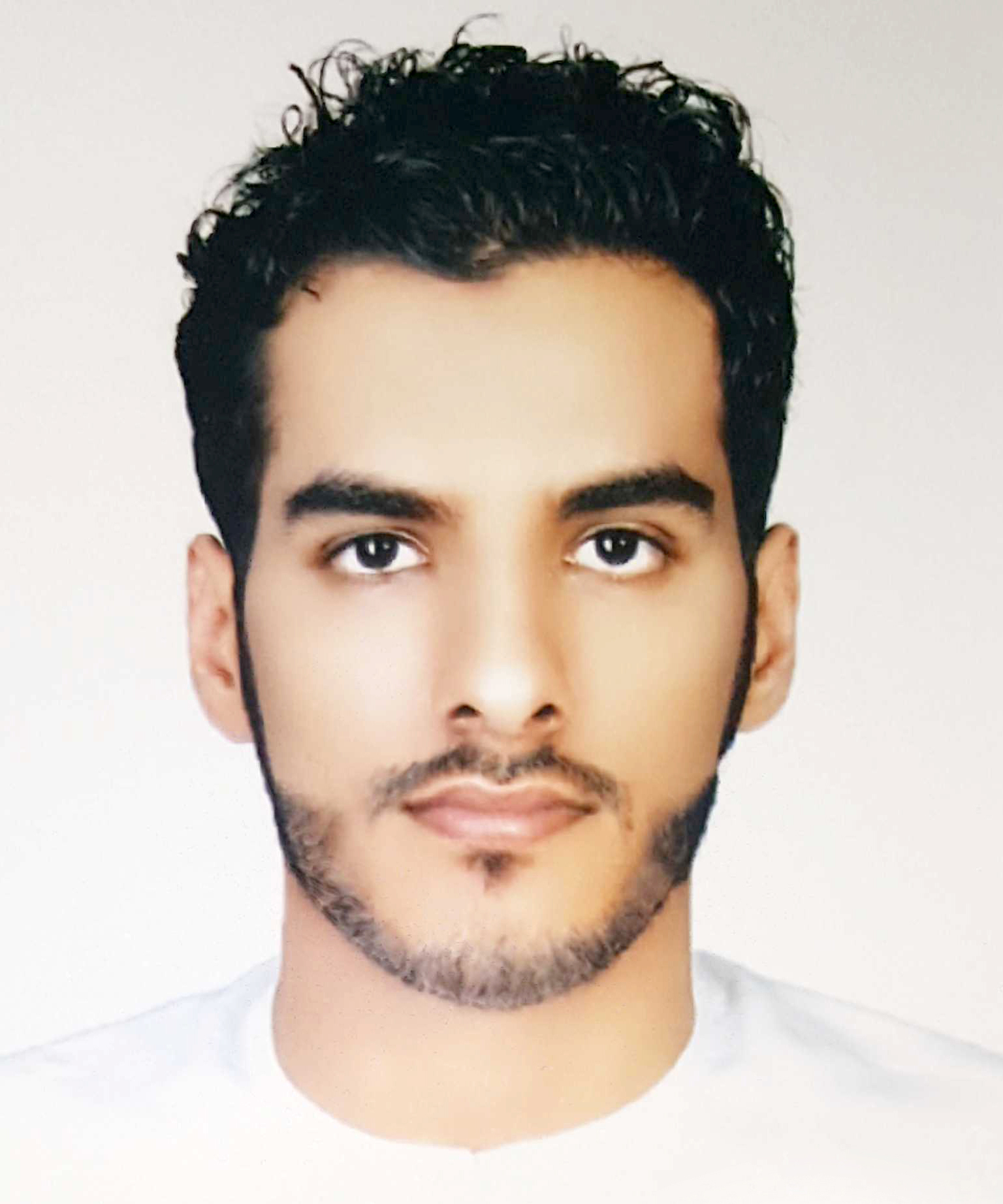}}]{Mohammed Bahutair}
is a PhD student in the School of Computer Science at  the University of Sydney, Australia. He received his bachelor degree in Computer Engineering from Ittihad University, UAE 2012 and his Masters degree in Computer Engineering from University of Sharjah, UAE 2015. His research interests are Machine Learning , Trust,  IoT and Big Data Mining.
\end{IEEEbiography}

\end{document}